\documentclass[conference]{IEEEtran}
\IEEEoverridecommandlockouts
\usepackage{url}
\usepackage{latexsym}
\usepackage{subcaption}
\usepackage{graphicx}
\usepackage{multirow}
\usepackage{array}
\newcolumntype{P}[1]{>{\centering\arraybackslash}m{#1}}
\newcolumntype{L}[1]{>{\raggedright\arraybackslash}m{#1}}
\newcommand{\ul}[1]{\underline{#1}}
\usepackage{enumitem}

\usepackage{cite}
\usepackage{amsmath,amssymb,amsfonts}
\usepackage{algorithmic}
\usepackage{graphicx}
\usepackage{textcomp}
\usepackage{xcolor}

\def\BibTeX{{\rm B\kern-.05em{\sc i\kern-.025em b}\kern-.08em
    T\kern-.1667em\lower.7ex\hbox{E}\kern-.125emX}}
\begin{document}

\title{Customizing Contextualized Language Models for Legal Document Reviews
}

\author{\IEEEauthorblockN{Shohreh Shaghaghian\IEEEauthorrefmark{1},
Luna (Yue) Feng\IEEEauthorrefmark{1}, Borna Jafarpour\IEEEauthorrefmark{1} and
Nicolai Pogrebnyakov\IEEEauthorrefmark{1}\IEEEauthorrefmark{2}}\\
\IEEEauthorrefmark{1} Center for AI and Cognitive Computing at Thomson Reuters, Canada\\
\IEEEauthorrefmark{2} Copenhagen Business School, Denmark\\
\\
%\IEEEauthorblockN{1\textsuperscript{st} Given Name Surname}
\IEEEauthorblockA{ \texttt{Emails: firstname.lastname@thomsonreuters.com}}}
%\textit{name of organization (of Aff.)}\\
%City, Country \\
%email address or ORCID}

\maketitle

\begin{abstract}
Inspired by the inductive transfer learning on computer vision,
many efforts have been made to train contextualized language models that boost the performance of natural language processing tasks. These models are mostly trained on large general-domain corpora such as news, books, or Wikipedia. Although these pre-trained generic language models well perceive the semantic and syntactic essence of a language structure, exploiting them in a real-world domain-specific scenario still needs some practical considerations to be taken into account such as token distribution shifts, inference time, memory, and their simultaneous proficiency in multiple tasks. In this paper, we focus on the legal domain and present how different language models trained on general-domain corpora can be best customized for multiple legal document reviewing tasks. We compare their efficiencies with respect to task performances and present practical considerations.
\end{abstract}

\section{Introduction}
Document review is a critical task for many law practitioners. Whether they intend to ensure that client filings comply with relevant regulations, update or re-purpose a brief for a trial motion,  negotiate or revise an agreement, examine a contract to avoid potential risks, or review client tax documents, they need to carefully inspect hundreds of pages of legal documents. Recent advancements in Natural Language Processing (NLP) have helped with the automation of the work-intensive and time-consuming review processes in many of these scenarios. Several requirements of the review process have been modeled as some common NLP tasks such as information retrieval, question answering, entity recognition, and text classification (e.g., \cite{sulea2017exploring}). However, certain characteristics of the legal domain applications cause some limitations for deploying these NLP methodologies. First, while electronic and online versions of many legal resources are available, the need for labelled data required for training supervised algorithms still needs resource-consuming annotation processes. Second, not only do legal texts  contain terms and phrases that have different semantics when used in a legal context, but also their syntax is different from general language texts. Recently, sequential transfer learning methods \cite{pan2009survey} have alleviated the first limitation by pre-training numeric representations on a large unlabelled text corpus using variants of language modelling (LM) and then adapting these representations to a supervised target task using fairly small amounts of labelled data. Attention-based Neural Network Language models (NNLM) pre-trained on large-scale text corpora and fine-tuned on a NLP task achieve the state-of-the-art performance in various tasks \cite{jing2019survey}. However, due to the second limitation, directly using the existing pre-trained models  may not be effective in the legal text processing tasks. These tasks may benefit from some customization of the language models on legal corpora.

In this paper, we specifically focus on adapting the state-of-the-art contextualized Transformer-based \cite{vaswani2017attention} language models to the legal domain and investigate their impact on the performance of several tasks in the review process of legal documents.  The contributions of this paper can be summarized as follows.
\begin{itemize}
    \item While \cite{elwany2019bert,chalkidis2020legal, zhang2020rapid} have fine-tuned the BERT language model \cite{devlin2019bert} on the legal domain, this study, to the best of our knowledge, is the first to provide an extensive comparison between several contextualized language models. Most importantly, unlike the existing works, it evaluates different aspects of the efficacy of these models which provides NLP practitioners in the legal domain with a more comprehensive understanding of practical pros and cons of deploying these models in real life applications and products.
\item Rather than experimenting with typical standalone NLP tasks, this work studies the impact of adaptation of the language models on real scenarios of a legal document review process. The downstream tasks studied in this paper are all based on the features of many of the existing legal document review products\footnote{\url{https://thomsonreuters.com/en/artificial-intelligence/document-analysis.html}

\url{https://kirasystems.com/solutions/} 

\url{https://www.ibm.com/case-studies/legalmation}} which are the result of hours of client engagement and user experience studies.
\end{itemize}
The rest of this paper is organized as follows. We review the existing studies related to our work in Section \ref{LitRev}. In Section \ref{Tasks}, we describe the legal document review scenarios we study. We briefly overview the language models we use in this paper in Section \ref{Models} and present the results of customizing and employing them in the document review tasks in Section \ref{Res}. The concluding remarks are made in Section \ref{Conclusion}.

\subsection{Related Works}\label{LitRev}

Large-scale pre-trained language models have proven to compete with or surpass state-of-the-art performance in many NLP tasks such as Named Entity Recognition (NER) and Question Answering (QA). There have been multiple attempts to do transfer learning and fine-tuning of these models on English NLP tasks \cite{peters2019tune}, \cite{yang2019end}.
However, these language models are trained on text corpora of general domains. For example, the Transformer-based language model, BERT\cite{devlin2019bert}, has been trained on Wikipedia and BooksCorpus \cite{zhu2015aligning}. The performance of these general language models are not yet fully investigated in more specific domains such as biomedical, finance or legal. To use these models for  domain-specific NLP tasks, one can repeat the pre-training process on a domain-specific corpus (from scratch or re-use the general-domain weights) or simply fine-tune the generic version for the domain-specific task.
\cite{lee2020biobert} has pre-trained generic BERT on multiple large-scale biomedical corpora and called it BioBERT. They show that while fine-tuned generic BERT competes with the state-of-the-art in several biomedical NLP tasks, BioBERT outperforms state-of-the-art models in biomedical NER, relation extraction and QA tasks. FinBERT \cite{araci2019finbert} authors pre-train and fine-tune BERT, ULMFit \cite{howard2018universal} and ELMO \cite{peters2018deep} for sentiment classification in the finance domain and evaluate the effects of multiple training strategies such as language model weight freezing, different training epochs, different corpus sizes, and layer-specific learning rates on the results. They show that pre-training and fine-tuning BERT leads to better results in financial sentiment analysis. BERT has also been pre-trained and fine-tuned on scientific data \cite{beltagy2019scibert} and it is shown that it provides superior results to fine-tuned generic BERT in multiple NLP tasks and sets new state-of-the-art results in some of them in the biomedical and computer science domains. 

One of the most relevant studies to our work is done by \cite{elwany2019bert} in which BERT is pre-trained on a proprietary legal contract corpus and fine-tuned for entity extraction task. Their results show that pre-trained BERT is faster to train for this downstream task and provides superior results. No other language models and NLP tasks have been investigated in this work. Another relevant work is \cite{chalkidis2020legal} in which the authors employ various pre-training and fine-tuning settings on BERT to fulfill the classification and NER tasks on legal documents. They report that the BERT language model pre-trained on legal corpora outperforms the generic BERT especially in the most challenging end-tasks (i.e., multi-label classification) where domain specific knowledge is more important. In this paper, we will have a more comprehensive investigation of the Transformer-based language models in the legal domain by pre-training and fine-tuning multiple Transformer-based language models for a variety of legal NLP tasks.

\section{Legal Review Tasks}\label{Tasks}
The formal and technical language of legal documents (i.e., legalese) is by nature often difficult and time consuming to read through. A practical facilitating tool is the one that can navigate the human reviewers to their points of interest in a single or multiple documents while asking for minimum inputs. Here, we have identified four main navigation scenarios in a legal document review process that can be facilitated by an automated tool. We elaborate on the main requirements that should be satisfied in each scenario and show how we model each task as an NLP problem. We explain the format of the labelled data we need as well as the learning technique we use to address each task. We eventually introduce a baseline algorithm for each task that can be used as a benchmark when evaluating the impact of using language models. Table \ref{ta:datasets} presents a summary of the details of each review task. Due to the complexities of legal texts, we define {\it snippet} as the unit of text that is more general than a grammatical sentence (see Section \ref{LLM} for an example of a legal text snippet). In our experiments, splitting the text into snippets is performed using a customized rule based system that relies only on text punctuation. 
\begin{table*}[h!]
\centering
\begin{tabular}{ |P{2.8cm}|P{3.2cm}|P{3cm}|P{2.8cm}|P{2.7cm}|}
 \hline
{\bf Review Task} & {\bf NLP Task} & {\bf ML Task} & {\bf Data Format*} & {\bf Baseline Algorithm}\\
 \hline
Information Navigation&  
Passage Retrieval& 
Binary Classification &
question snippet pairs& Siamese BiLSTM  \cite{tan2015lstm} \\
\hline
Fact Navigation&  
Named Entity Recognition & 
Sequence Labeling&
tokens &  \shortstack{2 Layer  BiLSTM \cite{graves2013speech} \\ (512 units) + dense layer}
\\
\hline
Comparative Navigation&  
Text Similarity&
Multi-class Classification&
snippet pairs & BiLSTM  \cite{graves2013speech}\\
\hline
Rule Navigation&
Sentiment Analysis&
Binary Classification&
\shortstack{snippets} &
XGBoost \cite{chen2016xgboost}\\
 \hline
\end{tabular}
\caption{Tasks in a legal document review scenario. *The format of the samples for the classifier.}
\label{ta:datasets}
\end{table*}

\subsection{Information Navigation}\label{qrl}
Navigating the users to the parts of the document where they can find the information required to answer their questions is an essential feature of any document reviewing tool. A typical legal practitioner is more comfortable with posing a question in a natural language rather than building proper search queries and keywords. These questions can be either factoid like {\it ``What is the termination date of the lease?"} or non-factoid such as {\it ``On what basis, can the parties terminate the lease?"}. Navigating the user to the answers of the non-factoid questions is equivalent to retrieving the potentially relevant text snippets and selecting the most promising ones.
This step can also be a prior step for factoid questions (see Section \ref{Fact Extraction}) to reduce the search space for finding a short text answer. 

We model this task as a classic passage retrieval problem with natural language queries and investigate the factoid questions for which the user is looking for an exact entity as another task. Given a question $q$ and a pool of candidate text snippets
$\{s_1, s_2, ..., s_N\}$, the goal
is to return the top $K$ best candidate snippets. Both supervised and unsupervised approaches have been proposed for this problem \cite{lai2018review}. We model this problem as a binary text classification problem where a pair of question and snippet $(q_i, s_i)$ receives a label $1$ if $s_i$ contains the answer to question $q_i$ and $0$ otherwise. We use the pairwise learning model proposed in \cite{tan2015lstm} as a baseline algorithm for this module. 

\subsection{Fact Navigation} \label{Fact Extraction}
As mentioned in Section \ref{qrl}, there are scenarios in which a legal professional needs to extract a pre-defined set of facts in a legal document. Examples of these scenarios are proofreading a single document to make sure that it contains the right set of facts or reviewing multiple documents to answer questions such as \textit{``What is the average of settlements awarded in a large set of court decision documents?"}.
We model this task as a sequence labeling problem in which each token will be assigned to a set of pre-defined classes using a multi-class classifier. For example, in a court decision document, the following set of classes may exist: \textit{Date of Argument}, \textit{Date of Decision}, \textit{Petitioner}, \textit{Judge}, \textit{Sought Damages} and \textit{Damages Awarded Monetary Values}. This task is similar to the Named Entity Recognition (NER) task. However, in legal review scenarios, there are some additional challenges: (1) Facts in some cases might span a large piece of text as opposed to a few tokens. For example, the commencement date in a real estate lease can be described as ``{\it Commencing on the later of: June 1,2018 or day on which Tenant enters into possession of the Premises (or on such earlier date that Tenant enters into possession of the Premises for the purpose of conducting its business therein)}". (2) This task is even more context-dependent. For example, in a general NER problem, the {\it Blackberry} token might refer to a company in one sentence and to a fruit in another sentence and they have to be distinguished. In a court decision ruling, the {\it Blackberry} token that refers to a company might take the role of {\it Plaintiff}, {\it Defendant} or has no role at all depending on the context.

\subsection{Comparative Navigation}
Comparing two pieces of text is an integral part of many legal document review scenarios. Whether the goal is to identify the differences between an amended regulation and its original version, to discover the discrepancies of regulations in different jurisdictions, or to investigate a legal agreement for potential deviations from an industry template, a law practitioner often needs to compare two or multiple legal documents \cite{alschner2019sense}. This task can be considered as a classic text similarity problem. However, one of the main challenges in legal text comparison is that the same legal concept may be expressed in very different ways, complicating the application of lexical matching approaches such as tf-idf. For example, ``{\it Broker is entitled to a commission from the vendor}" and ``{\it The seller shall pay the agent a fee}" carry the same legal meaning, yet have practically no words in common. This makes this task well-suited for language modeling.

The problem is formulated as follows. We are given one document consisting of $N_r$ text snippets, another document of $N_t$ snippets and a set of labels $L = \{l_1, l_2, \dots \}$. The labels can be binary, such as $L = \{match, \textit{no match}\}$ or multiclass, e.g., $L = \{\textit{match}, \textit{partial match}, \textit{no match}\}$. We want to assign a label $l^{i,j}\in L$ to any snippet pair $(s_r^i, s_t^j)$ where $i \in \{1,\dots N_r\}$ and $j \in \{1,\dots N_t\}$. In the experiments of this paper, we focus on the binary label scenarios.

\subsection{Rule Navigation}
One of the main purposes of reviewing a legal document by a legal expert is to identify the indispensable rules imposed by deontic modalities. In legal documents, modalities are ubiquitously used  for representing legal knowledge in the form of obligations, prohibitions and permission \cite{neill2017classifying}. In practice, legal professionals identify deontic modalities by referencing the modal verbs specially ``{\it would}", ``{\it should}" and ``{\it shall}" that express obligatory and permissive statements. However, solely depending on the modal verbs to automatically identify the deontic modalities is error prone due to three main reasons: (1) It is difficult to quantify the semantic range between modalities \cite{verstraete2005scalar}. For example, the person who is {\it allowed} or {\it obliged} to do something carries different sentiment; (2) Modal verbs can have more than one function that may not indicate a deontic modality. For instance, in the sentence ``{\it Licence Agreements shall mean collectively, the Trademark Licence and the Technology Licence}", the modal verb ``{\it shall}" does not indicate any deontic modality; (3) The misuse of modal verbs in the documents brings in another complexity to disambiguate from the real deontic modalities. Therefore, context is important for interpreting what the modal verbs are meant and whether deontic modalities are presented. This task can be considered as a sentiment analysis problem.

To simplify the problem, we model it as a binary classification problem aiming to identify text snippets that contain obligations, particularly positive duties from the full document and consider the rest of the snippets having no deontic modalities. Formally, given a document with text snippets $\{s_{1}, s_{2}, ..., s_{N}\}$, the binary classifier will label each snippet based on whether it imposes a positive duty for a party or not.

\section{Language Models}\label{Models}
In this section, we first briefly review the four Transformer-based language models we study in this work. Note that the language models we investigate in this paper are only made of the encoder of the transformer architecture. Then, we elaborate on how we adapt them to the legal domain. Table \ref{table_LMs} provides a summary of different characteristics of the language models studied in this work. 

\begin{table*}[h!]
\centering
\begin{tabular}{ |P{1.7cm}||P{1.9cm}|P{1.8cm}|P{2.5cm}|P{1.8cm}|P{3.1cm}|}
 \hline
{\bf Model Name}& {\bf Number of Parameters}& {\bf Model Size on Disk} & {\bf Number of Hidden Layers} &{\bf Number of Tokens}&{\bf Training Time*} (hours / epoch ) \\
 \hline
 BERT &   110 M &416 MB&12&28,996& 7.12 \\
  \hline
 DistilBERT &66 M&252 MB & 6&28,996&7.02 \\
 \hline
  RoBERTa   &125 M&501 MB&12&50,265&7.87\\
  \hline
  ALBERT   &12 M&45 MB& 12&30,000&7.03\\
  \hline
\end{tabular}
\caption{Comparison of some features of language models. The maximum window length is 512 and the training task is MLM for all four models.  *Training times for BERT and DistilBERT are averaged over different tokenization and weight initialization methods.}
\label{table_LMs}
\end{table*}

\subsection{BERT}
Proposed by \cite{devlin2019bert}, Bidirectional Encoder Representations from Transformers (BERT) aims to capture the context from both left-to-right and right-to-left directions when learning a representation for a text snippet. The input tokenization of BERT is WordPiece \cite{wu2016google} and its architecture is a multi-layer bidirectional Transformer encoder. Transformer \cite{vaswani2017attention} is a sequence transduction model in which the recurrent layers are replaced with multi-headed self attention (i.e., simultaneous attention to different parts of the sequence). Recurrent Neural Networks (RNNs) need to be unwrapped in the order of the input. This constraint makes training and inference processes time-consuming for long sequences specially in the encoder-decoder architectures. By replacing the recurrent connections, more parallelism is achieved since each node’s output can be calculated solely based on previous layer’s output (as opposed to RNN nodes that need the output of the previous node in the same layer). Due to the lack of recurrence, positional embeddings are trained for each of the input positions. 
%The encoder and decoder in this architecture are the same except for an extra attention mechanism in the decoder for the encoder outputs. 
BERT's language model is simultaneously trained with two tasks of Masked Language Modeling (MLM) and Next Sentence Prediction (NSP). However, ablation studies in \cite{devlin2019bert} and subsequent publications \cite{liu2019roberta} show that NSP has minimal contribution to downstream tasks' performance metrics.  In this work, we use bert-base-cased trained by Google on English Wikipedia and Toronto Book Corpus \cite{zhu2015aligning} as the general-domain version of the BERT language model.

\subsection{DistilBERT}
DistilBERT \cite{sanh2019distilbert} is a smaller and faster version of BERT which is reported to achieve 97\% of BERT's performance on GLUE \cite{wang2018glue} while reducing the size by 40\% and the inference time by 60\%. In order to train this language model, a technique called knowledge distillation, also referred to as teacher-student learning, is incorporated which is originally proposed in \cite{hinton2015distilling} and \cite{bucilua2006model}. The goal of this technique is to compress a large model (i.e., the teacher) into a smaller model (i.e., the student) that can reproduce the behaviour of the larger model. 
In this teacher-student setup, the student network is trained by matching the full output distribution of the teacher network rather than by maximizing the probability of the correct class. Therefore, instead of training with a cross-entropy over the hard targets (i.e., one-hot encoding of the correct class), the knowledge from the teacher to the student can be transferred with a cross-entropy loss function over the predicted probabilities of the teacher i.e., $L_{ce} = -\sum_{i}t_{i} * log(s_{i})$
where $t_i$s and $s_i$s are the probabilities respectively estimated by the teacher and the student. The final training objective is a linear combination of the distillation loss $L_{ce}$ and the supervised training loss which is the MLM loss. In the process of distilling the BERT model, the student network uses the same architecture as BERT. The token-type embeddings and the pooler are removed while the number of layers is reduced by a factor of 2. As for the general-domain version of DistilBERT, we use distilbert-base-cased.

\subsection{RoBERTa}
RoBERTa is a release of the BERT architecture that optimized the training regime and used a larger dataset for pre-training \cite{liu2019roberta}. Its release was prompted by the observation that the original BERT model \cite{devlin2019bert} followed a suboptimal training procedure. Specifically, the changes made in RoBERTa include (i) training only with the MLM objective (removing BERT's NSP objective), (ii) a larger batch size of 8,000 (compared to 256 in BERT), (iii) a higher learning rate, peaking at 4e-4 for the large model and 6e-4 for the base model (vs. 1e-4 for BERT), (iv) byte-level vocabulary with 50K subword tokens (unlike BERT's character-level vocabulary of 30K tokens), which added 15M parameters to the model for the base version compared to BERT, and (v) a larger training dataset of 160 GB (compared to 13 GB in BERT), including BERT's training set and an additional 76-GB set compiled from news articles. With these modifications, RoBERTa is reported to outperform other language models trained after BERT's release such as XLNet. Here, we use roberta-base pretrained by Facebook as the general-domain version of the model. 

\subsection{ALBERT}
Larger language models are shown to lead to better accuracy in many NLP downstream tasks. However, the size of these models are constrained by computation cost, training time and GPU memory. To address these issues, \cite{lan2019albert} propose A Lite BERT (ALBERT)
language model which has significantly fewer parameters than the original BERT model by employing the following two techniques.
\begin{itemize}[leftmargin=*,topsep=1pt,itemsep=0.5pt,partopsep=4pt, parsep=4pt]
\item {\bf Factorized embedding parameterization}: BERT with vocabulary size of $V$ and embedding size of $E$ has $VE$  parameters in the embedding layer which are sparsely updated during training. Instead of using one-hot encodings to be projected to the embedding layers directly, ALBERT first projects them into a smaller space of size $S$ and then into the embedding space of size $E$. Therefore, ALBERT embedding layer size is $VS +SE$  which can be significantly smaller than BERT's embedding parameters if $S<<E$. Also, parameters of the embedding layer are less sparse and are updated more frequently.
\item {\bf Weight sharing}: All weights are shared between all layers of ALBERT. 
ALBERT also uses a slightly different loss function compared to BERT. In this new loss function, BERT's NSP task is replaced with the sentence order prediction (SOP) task. In SOP task, positive examples are the same as NSP but negative examples are the same two sentences with their order reversed. Auhtors speculated that NSP is a much simpler task than MLM and replacing it with SOP will help the model to better learn natural language coherence.
\end{itemize}
\subsection{Legal Domain Language Models}\label{LLM}
Before being able to propose adjustment to the pre-trained language models, we need to understand how the language of a document written in legalese can be different from plain English. In the rest of this paper, we focus on the language of legal agreements as the legal domain language and use a subset of 9,000 legal agreements of the publicly available corpus of US Securities and Exchange Commission (SEC)\footnote{\url{https://www.sec.gov/edgar.shtml}} as our domain specific corpus. For all the four language models explained above, we use the HuggingFace Transformers library \cite{wolf2019huggingface} for training of the language models and for the downstream tasks. There are two main distinguishing features in the language of legal documents. 
\begin{itemize}[leftmargin=*,topsep=1pt,itemsep=1pt,partopsep=4pt, parsep=4pt]
\item {\bf Domain Specific Terms} There are some terms and phrases that are unique to law such as ``\textit{fee simple}" or “\textit{novation}”. However, regular words like “\textit{party}” or “\textit{title}” may have different semantics when used in a legal context compared to a general context. In addition, many old words and phrases such as “\textit{herein}”, “\textit{hereto}”, “\textit{hereby}” and “\textit{heretofore}”, as well as some non-English words like “\textit{estoppel}” or “\textit{habeas corpus}” are often employed in legalese. The models trained on general-domain corpora have either never seen these terms or have captured their general semantics. 
\item {\bf Syntactic Structure} A single sentence written in legalese can have very long and complex construction. For example, the language models pre-trained on general-domain corpora have barely seen the complex syntactic structures like this:

“\textit{In the event of any sale of such interest or transfer of such rights and upon the assumption, in writing, of the obligations of Landlord under this Lease by such assignee or transferee, Landlord herein named (and in case of any subsequent transfer, the then assignor) shall be automatically freed and relieved from and after the date of such transfer of all liability in respect of the performance of any of Landlord’s covenants and agreements thereafter accruing, and such transferee shall thereafter be automatically bound by all of such covenants and agreements, subject, however, to the terms of this Lease; it being intended that Landlord’s covenants and agreements shall be binding on Landlord, its successors and assigns, only during and in respect of their successive periods of such ownership\footnote{\url{https://www.sec.gov/Archives/edgar/data/1600132/000110465915078036/a15-18062_1ex10d1.htm}}.}”
\end{itemize}

These examples confirm the need for the language models that can capture the specific syntactic and semantic features of the legal domain. Table \ref{datastats} shows the probability distribution of number of words in a sentence for a general domain English corpus \cite{guo2020wiki} and our legal corpus. We see that, using the same sentence splitting rule, there are on average 73 more words in a legal sentence. 

\begin{table*}[h!]
\centering
\begin{tabular}{ |P{2cm}|P{1.4cm}|P{2cm}|P{5cm}|P{6cm}|}
 \hline
{\bf Corpus Domain} & {\bf Tokenization}& {\bf Average Tokens per Sentence}& {\bf Sentence Length (Number of Tokens) Histogram} & {\bf Tokenization Examples*}  \\
\hline
  General & Words   &27 &\includegraphics[width=\linewidth]{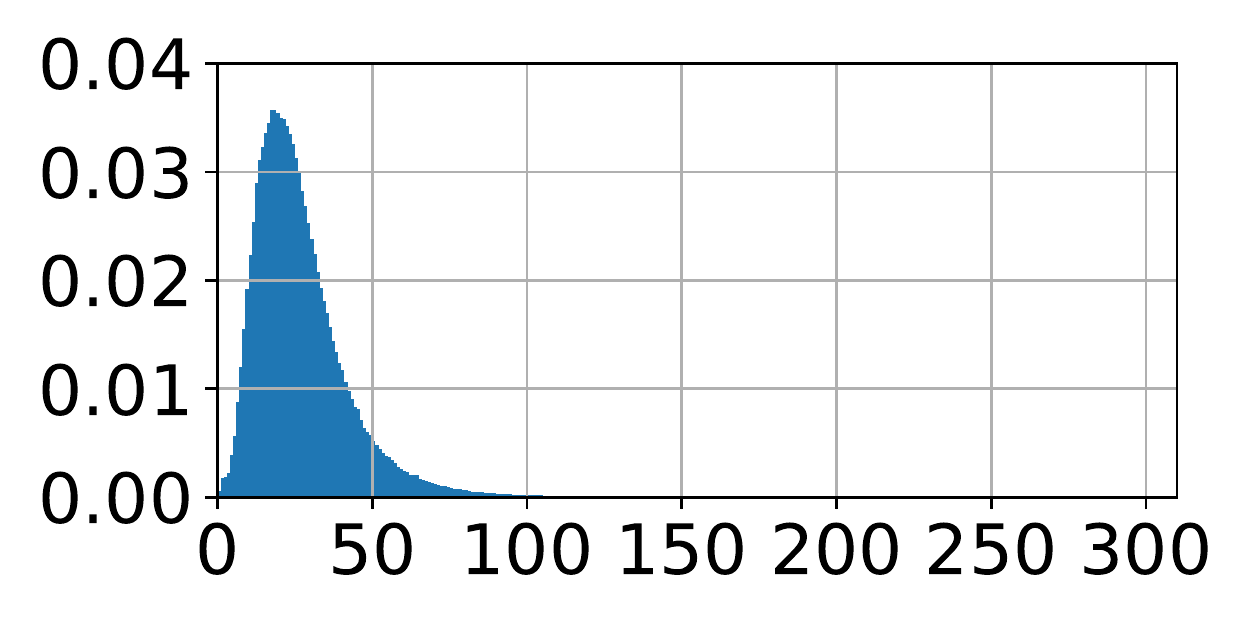}& ['attorney-in-fact', 'injunctive', 'self-insurance', 'contingency', 'Condominium']\\
 \hline
 Legal &  Words &   100 &\includegraphics[width=\linewidth]{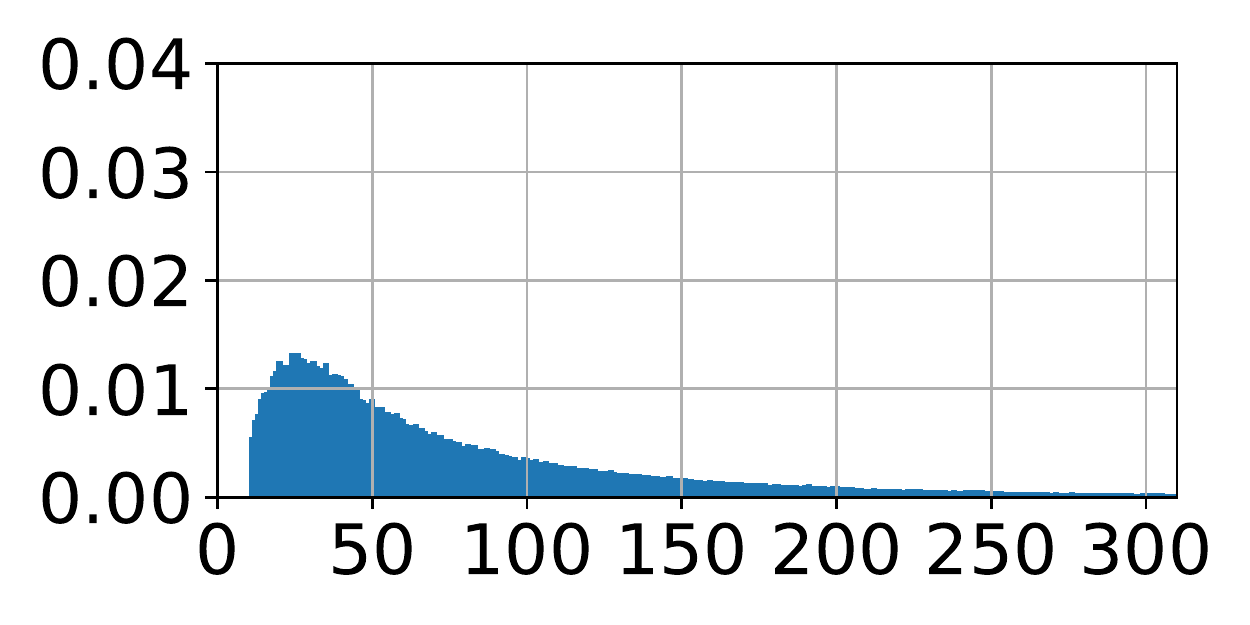} &['attorney-in-fact', 'injunctive', 'self-insurance', 'contingency', 'Condominium'] \\
  \hline
 Legal & General Tokens & 142 &\includegraphics[width= \linewidth]{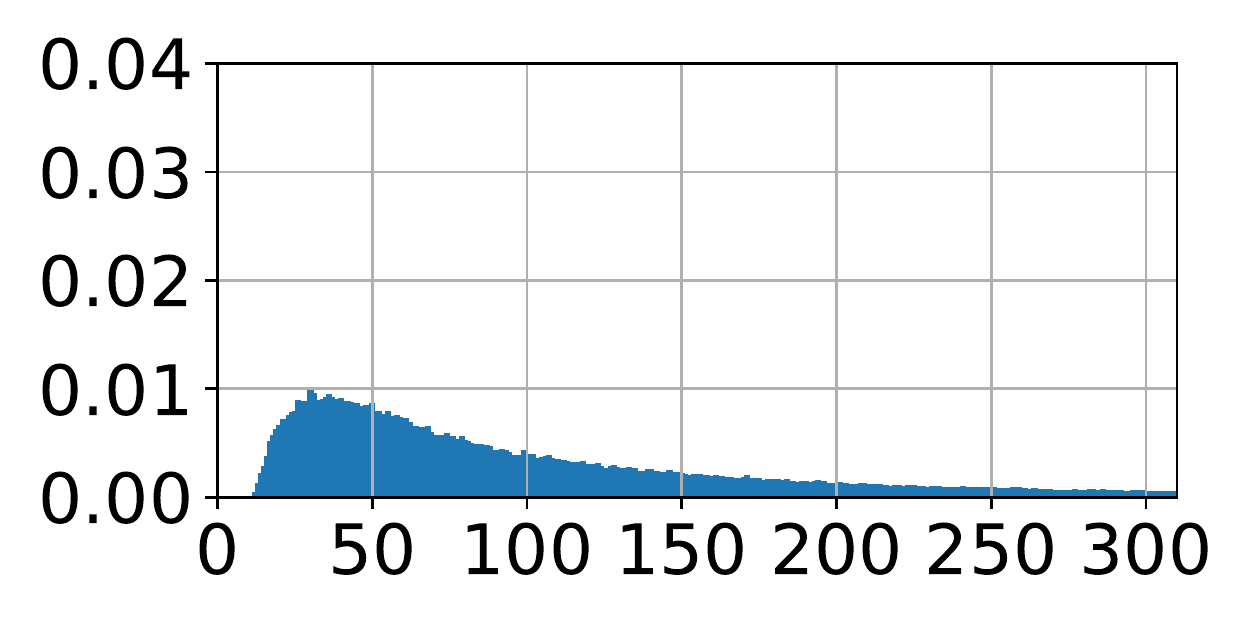} & ['attorney', '-', 'in', '-', 'fact', 'in', '\#\#junct', '\#\#ive', 'self', '-', 'insurance', 'con', '\#\#ting', '\#\#ency', 'Con', '\#\#dom', '\#\#ini', '\#\#um'] \\
 \hline
Legal &   Legal Tokens   & 142 &\includegraphics[width=\linewidth]{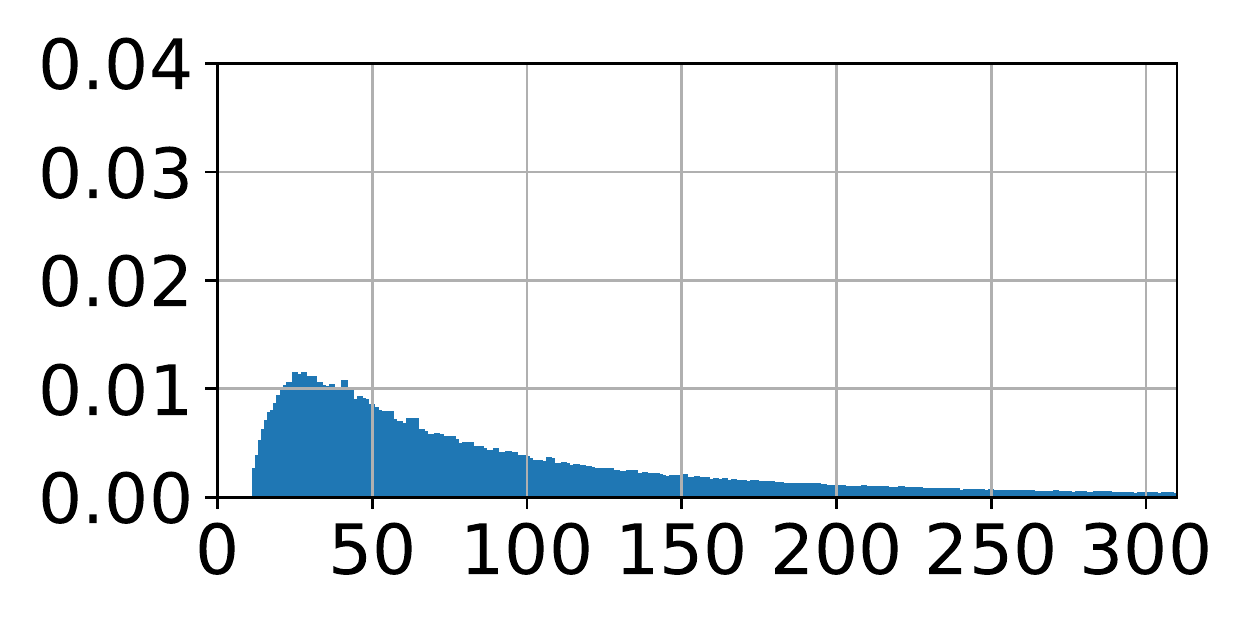}&['attorney', '-', 'in', '-', 'fact', 'injunctive', 'self', '-', 'insurance', 'contingency', 'Condominium'] \\
  \hline
 Legal &  Hybrid Tokens   & 120 &\includegraphics[width=\linewidth]{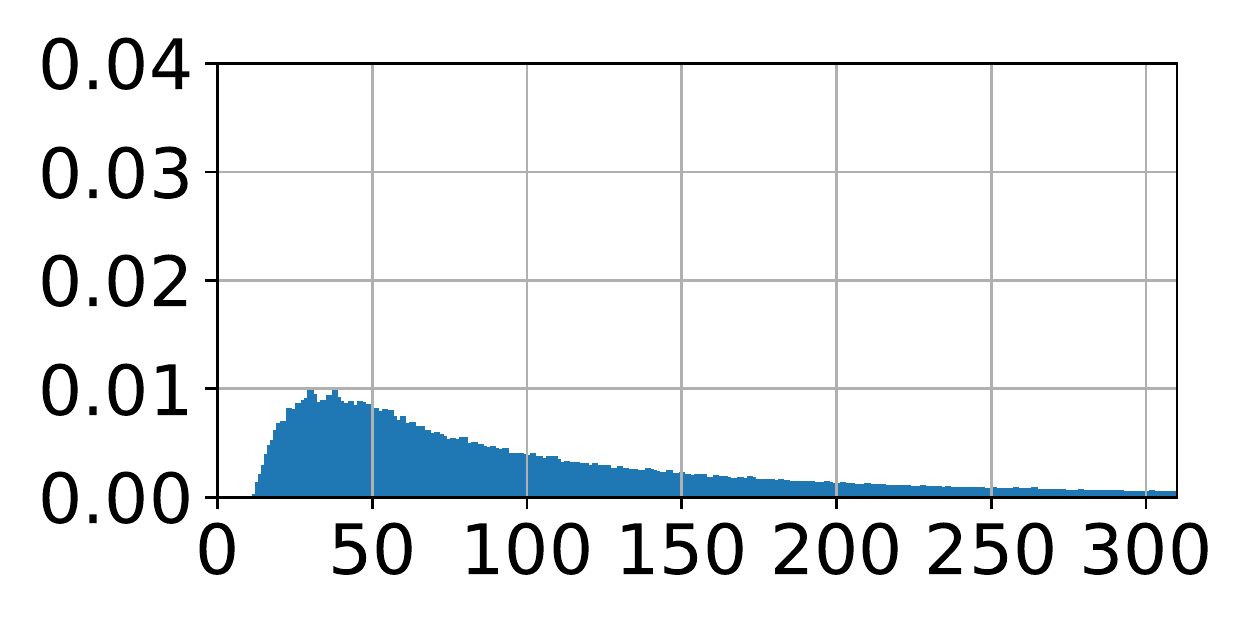}& ['attorney-in-fact', 'injunctive', 'self-insurance', 'contingency', 'Con', '\#\#dom', '\#\#ini', '\#\#um']
 \\
  \hline
\end{tabular}
\caption{Comparison of distributions of number of tokens per sentence for different tokenization approaches in different domains, with examples. *Examples show how each tokenization approach tokenizes the sequence ``attorney-in-fact injunctive self-insurance contingency Condominium"}
\label{datastats}
\end{table*}

\section{Experiments}\label{Res}
While all the experiments of this paper have focused on the domain of commercial real estate agreements, the methodologies in defining the tasks, building the datasets, and customizing the models can be generalized for similar review scenarios in other domains and other types of legal documents.
There are three main questions we try to answer in this paper: (1) What are different approaches to adapt a contextualized language model to the legal domain?; (2) How do different language model architectures perform for different review tasks?; (3) How to choose a model if the computational resources are limited?

\begin{itemize}[leftmargin=*,topsep=1pt,itemsep=1pt,partopsep=4pt, parsep=4pt]
\item {\bf What are different approaches to adapt a contextualized language model to the legal domain?}
\end{itemize}

 In order to answer the first question, we study the impact of two main factors on training of language models: tokenization and initial weights. We use Sentencepiece\footnote{\url{https://github.com/google/sentencepiece}} on our legal corpus to generate the same number of cased tokens as in bert-base-cased's token set (See Table \ref{table_LMs}). We call these domain-specific tokens Legal Tokens while we refer to bert-base-cased's original tokens as General Tokens. Only $~36\%$ of tokens are common between Legal Tokens and General Tokens. While tokens like {\it attorney}, {\it lease},  and {\it liability} are common unbroken tokens in these two sets, other more domain-specific tokens such as {\it lessor}, {\it lessee}, and {\it memorandum} only exist in the legal set. We also use a hybrid version in which we add the 500 most frequent words in our legal corpus that do not exist as an independent unbroken token in the set of General Tokens. This set is referred to as Hybrid Tokens. We limited the number of added tokens to only 500 because of overhead it adds to the size the embedding layer and therefore the training time.
 
 Table \ref{datastats} shows the probability distribution functions of number of tokens in the sentences of our legal corpus using these three tokenization approaches. As we see in the table, by switching from General Tokens to Legal Tokens, we do not see a significant change in the distribution of the number of tokens in sentences. That is due to the fact that the number of generated tokens are the same in these two tokenizations. However, the way a single sequence has been tokenized is different is these two approaches. On the other hand, by adding the 500 most frequent tokens in the hybrid tokenization approach, the average number of tokens in sentences decreases. By comparing the tokenization examples, we realize that the word {\it contingency} is among the top 500 most frequent words of the legal corpus and Sentencepiece also captures it as a single token in the legal corpus. However, it is probably not a very frequently used token in the general corpus of bert-base-cased therefore it is broken into sub-tokens using the general tokenization approach. The word {\it Condominium} is not among the top 500 tokens of the legal corpus, therefore the tokenization based on Hybrid Tokens breaks it into sub-tokens, but Sentencepiece still captures it as a single unbroken token when creating Legal Tokens. Finally, the hyphenated compound words {\it self-insurance} and {\it attorney-in-fact} are among the top 500 most frequent words in the legal corpus, but based on rules of Sentencepiece, they are broken in both General Tokens and Legal Tokens.

\begin{figure*}[htp]
  \centering
  \includegraphics[width=0.85\linewidth]{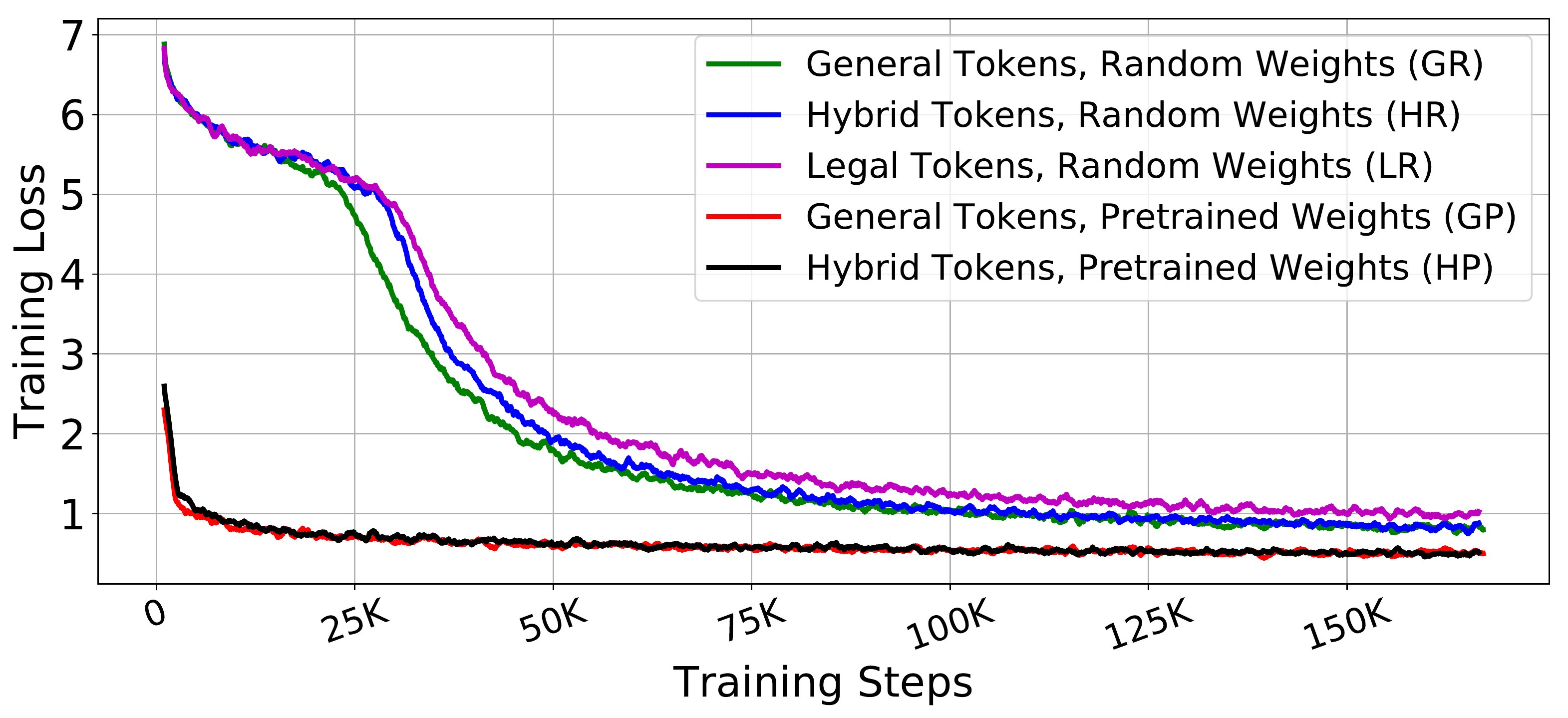}
\caption{Training loss through pre-training of BERT with different tokenization and initial weights}
\label{fig:lm_losses}
\end{figure*}

 Based on the two factors of tokenization and initial weights, we train five different versions of the BERT language model. For general and hybrid tokenization approaches (in which the majority of tokens are general-domain tokens), we start the training both from the general-domain model weights published with the original papers (i.e., pre-trained initial weights) and from scratch (i.e., random initial weights). Figure \ref{fig:lm_losses} shows the moving average of the MLM training loss through 10 epochs of training the BERT language model on a p3.8xlarge AWS instance (i.e., 4 Tesla V100 GPUs with 64 GB memory) with batch size of 32 samples with max lengths of 512 tokens. We used Adam optimizaer with learning rate of $3\times 10^{-5}$. As we see in the figure, the training loss saturates much faster when starting from the pre-trained weights. Also, comparing GR, HR, and LR models, we see that adding more domain-specific tokens delays the saturation in loss. 

\begin{itemize}[leftmargin=*,topsep=1pt,itemsep=1pt,partopsep=4pt, parsep=4pt]
\item {\bf How do different language model architectures perform for different review tasks?}
\end{itemize}
In order to answer the second question, we compare the result of BERT, DistilBERT, RoBERTa, and ALBERT on our downstream document review tasks. We use the aforementioned five versions of BERT as the teacher to customize the DistilBERT language models. However, for RoBERTa and ALBERT we only compare the base version (i.e., without any customization for the legal domain) with only one customized version using general tokens and starting from pre-trained weights.

As mentioned in Section \ref{Models}, we use publicly available legal agreements and use some human-annotated labels for each of the tasks described in Section \ref{Tasks}. 
The labelled dataset for all of the four legal review tasks is created by legal practitioners with detailed instructions of the annotating process. However, due to the complexity of the annotating tasks and limited resources (for example, it took a legal expert annotator on average 4 hours to annotate the snippets related to the passage retrieval task in one document), each instance is only labelled by one human which prevents us from computing the inter-annotator agreement score. We remedied that by providing comprehensive instructions to annotators, engaging annotators in pre-task training, as well as performing pilot annotations and reviewing their results with a senior law practitioner before proceeding with full annotation. Table \ref{data_samples} shows examples of data samples for each of the four tasks.
\begin{table*}[t!]
\centering
\begin{tabular}{ |P{1.5cm}|P{14.5cm}|P{1cm}|}
 \hline
 {\bf NLP Task} & {\bf Examples} &\bf{Label}\\
 \hline
\multirow{2}{*}{\shortstack{\bf Passage\\ \bf Retrieval}} & 
\multicolumn{1}{L{14.5cm}|}{ \textbf{Question:} Does the tenant have the right to challenge tax assessments? 
\break\break
\textbf{Snippet:} If the Premises separately assessed, Tenant shall have the right, by appropriate proceedings, to protest or contest in good faith any assessment or reassessment of Taxes , any special assessment, or the validity of any Taxes or of any change in assessment or tax rate; provided, however, that prior to any such challenges must either (a) pay the taxes alleged to be due in their entirety and seek a refund from the appropriate authority, or (b) post bond in an amount sufficient to insure full payment of the Taxes.\break }& 1
\\
\cline{2-3}
&
\multicolumn{1}{L{14.5cm}|}{\textbf{Question: }Does the tenant have the right to challenge tax assessments?
\break
\textbf{Snippet:} Landlord also shall provide Tenant with a copy of the applicable Tax bill or Tax statement from the taxing authority.\break}& 0
\\
\hline
\multirow{2}{*}{\shortstack{\bf Text \\ \bf Similarity}} &

\multicolumn{1}{L{14.5cm}|}{\break {\bf Snippet1}: The exercise of any remedy by either party shall not be deemed an election of remedies or preclude that party from exercising any other remedies in the future, except as expressly set forth herein. 
\break\break
{\bf Snippet2}: Either party's acceptance of monies under this Lease following a Default by the other shall not waive such party's rights regarding such Default.\break}&1
\\
\cline{2-3}
& 
\multicolumn{1}{L{14.5cm}|}{\textbf{Snippet1}. Provided Tenant has performed all its obligations, Tenant shall peaceably and quietly hold and enjoy the Premises for the Term, subject to the provisions of this Lease.
\break \break
\textbf{Snippet2}. If either party elects to terminate this Lease as provided in this Section, this Lease shall terminate on the date which is 30 days following the date of the notice of termination.\break}&0
\\
\hline
\multirow{2}{*}{\shortstack{{\bf Named Entity}\\{\bf Recognition}}} &
\multicolumn{1}{L{14.5cm}|}{MINUTE ORDER IN CHAMBERS - JUDGE ORDER PERMITTING PLAINTIFF TO FILE ADDITIONAL SUPPLEMENTAL BRIEFS REGARDING 35 USC 101 by Judge George H. Wu. {\bf BlackBerry} is permitted to file an additional supplemental brief by April 23, 2020. \break}& Plaintiff
\\
\cline{2-3}
 &
\multicolumn{1}{L{14.5cm}|}{ Tenant will pay a security deposit of Nineteen Million Seven Hundred and Fifty Thousand Dollars ( \$ 19,750,000 ) ( {\bf payable in cash or , as and to the extent set forth in Section 3.7.1 , in the form of a letter of credit reasonably acceptable to Landlord}  ) ( the ``Security Deposit"). \break} & Form of Security Deposit
\\
\hline
\multirow{2}{*}{\shortstack{\bf Sentiment\\ \bf Analysis}}&

\multicolumn{1}{L{14.5cm}|}{{\bf Snippet:} If the Landlord so requires the Principal Rent shall be paid directly to the Landlord's bankers by bankers standing order.\break} &1\\
\cline{2-3}
&
\multicolumn{1}{L{14.5cm}|}{{\bf Snippet:} "Prescribed Rate" shall mean such comparable rate of interest as the Landlord reasonably determines.\break}&0
\\
\hline
\end{tabular}
\caption{Examples of labeled data for each of the 4 review tasks. }
\label{data_samples}
\end{table*}

For the passage retrieval task (i.e., information navigation), we use 106 documents and have human legal experts annotate the answers to 31 questions in each one. In a single document, each question may have zero, one, or multiple (consecutive or non-consecutive) snippets as its answers. For each question, negative samples are generated by uniformly sampling 10 snippets from the snippets of the document that are not annotated as the answer to that question. Overall, we use a set of approximately 81,000 question-snippet pairs and compare the impact that different models have on the passage retrieval task based on the F1-score of the classifier.

In the text similarity task (i.e., comparative navigation), we have used the same documents used for the passage retrieval task and sampled pairs of text snippets from these documents. In order to reduce the search space for finding the matches, we limit ourselves to the pieces of text that have been annotated as the answer to the same question in the retrieval task. We have then asked human legal experts to label each pair as a match or not. The snippets of a relevant pair are the ones that address the same legal topic and a specific set of points within that topic, for example, a specific set of responsibilities of a party to the agreement. We build a dataset of approximately 1,500 snippet pairs. The distribution of labels in the data for this task is 40\% in the positive class to 60\% in the negative class. We present the F1-score as the metric for evaluating different models.

\begin{figure}[htp]
  \centering
  \includegraphics[width=1.03\linewidth]{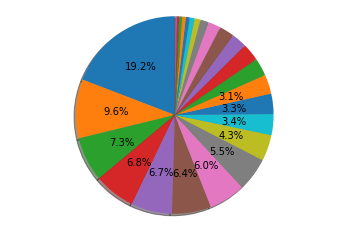}
\caption{Distribution of named entities' tokens among our 26 classes, ranging from 19.2\% to 0.1\% of the whole named entity tokens}
\label{fig:ner_dist}
\end{figure}

In the entity recognition task (i.e., fact navigation), we use data from 200 documents fully annotated by human experts. This dataset contains of approximately 5,000 snippets whose tokens are annotated with at least one of the 26 entities of interest. A text snippet might contain multiple entities. Only 3\% of the sentences contain an entity and only 1\% of tokens are part of an entity. We use 20,000 randomly sampled snippets as negative samples and report token level micro average F1-score. We use token level F1-score as it provides a more fine-grained evaluation compared to entity level evaluation. Figure
\ref{fig:ner_dist} shows the distribution of tokens that are annotated as part of an entity among our 26 classes. 

For the sentiment analysis task (i.e., rule navigation), we have the human legal experts assign labels to approximately 3,000 text snippets randomly selected from our document corpus. If the snippet contains a positive duty statement, a label 1 is assigned to it. Otherwise, the label is 0. After assigning human labels, 24\% of samples are labeled as 1. We present the F1-score as the metric for comparing various language models.

Table \ref{table_performance_results} reports the performance metrics of document review tasks by adding a linear layer as the sequence  (or token in case of NER) classification heads on top of the pooled (or hidden-states) output of different versions of contextualized language models. The base version of each model corresponds to the general-domain language models published with the original papers without any adjustment for the legal domain. 
For all four tasks, we split the datasets into training, validation, and test sets with respectively 80\%, 10\% and 10\% ratios. The reported results are the average over 3 different random splits. The standard deviations are also reported in the parentheses. In each task, the same stopping criteria (e.g, early stopping based on the validation loss) is used for all models. 

\begin{table*}[h!]
\centering
\begin{tabular}{|P{1.4cm}||P{1cm}|P{1.1cm}|P{2cm}|P{2.3cm}|P{2.1cm}|P{2.3cm}|P{2.4cm}|}
\hline
{\bf Model}  &
{\bf Version} &{\bf Tokens} &{\bf Initial Weights} &
{\bf Passage Retrieval} & {\bf Text Similarity} & {\bf Entity Recognition} &{\bf Sentiment Analysis} \\ \hline
\multirow{6}{*}{\shortstack{{\bf BERT}}}
& base &General & --     & 0.83  ($\pm$ .05)    &0.73 ($\pm$ .04)   &\ul{0.46} ($\pm$.01)    &0.84 ($\pm$ .01)        \\
& GR &General &Random     & 0.84 ($\pm$ .05)     &0.79 ($\pm$ .05)    &0.21 ($\pm$.00)     &0.83 ($\pm$ .02)        \\
&HR &Hybrid &Random     &  0.84 ($\pm$ .04)    &0.78 ($\pm$ .04)    &0.33 ($\pm$ .01)     &  0.82 ($\pm$ .02)          \\
&LR &Legal &Random       & 0.85 ($\pm$ .03)     &0.79 ($\pm$ .02)    &      0.32 ($\pm$.01)  & 0.85 ($\pm$ .04)      \\
&GP &General &Pre-trained & \ul{\bf 0.86} ($\pm$ .05)     &    \ul{\bf0.82} ($\pm$ .02)         & 0.42 ($\pm$.03)       & \ul{\bf0.92} ($\pm$ .01)        \\
&HP &Hybrid &Pre-trained   &0.85 ($\pm$ .05)      &    0.79 ($\pm$ .04)         & 0.45 ($\pm$.01)        &  0.89 ($\pm$ .02)      \\
\hline 
\multirow{6}{*}{{\bf DistilBERT}}
&base &General &--        &0.81 ($\pm$ .06)     &    \ul{0.76} ($\pm$ .03)         &     \ul{\bf 0.48} ($\pm$ .03)       & 0.88 ($\pm$ .01)         \\ 
&GR &General &Random      &   0.82 ($\pm$ .04)      &    0.75 ($\pm$ .04)        &      0.42 ($\pm$.01)      & 0.87 ($\pm$ .03)        \\
&HR &Hybrid &Random       & \ul{0.83}   ($\pm$ .04)    &    0.73 ($\pm$ .05)      &       0.37 ($\pm$.04)      &     0.87 ($\pm$ .02)    \\
&LR &Legal &Random      &  \ul{0.83} ($\pm$ .05)       & \ul{0.76} ($\pm$ .03)          &  0.32 ($\pm$ .01)           &     \ul{0.91} ($\pm$ .02)    \\
&GP &General &Pre-trained  & 0.81  ($\pm$ .04)       &    0.74 ($\pm$ .04)        &      0.41 ($\pm$.03)       & 0.89 ($\pm$ .01)        \\
&HP &Hybrid &Pre-trained   & 0.82  ($\pm$ .04)       &    0.74 ($\pm$ .03)         &       0.34 ($\pm$.03)      &   0.88 ($\pm$ .03)     \\
\hline
\multirow{2}{*}{{\bf RoBERTa}}
&base &General &--           &\ul{0.84} ($\pm$ .05) &      \ul{0.79} ($\pm$ .01)       &      \ul{0.38} ($\pm$.00)       & 0.89 ($\pm$ .03)        \\
&GP &General &Pre-trained      & \ul{0.84} ($\pm$ .05)         &    0.75 ($\pm$ .02)         &       \ul{0.36} ($\pm$.02)      & \ul{0.91} ($\pm$ .03)            \\
\hline
\multirow{2}{*}{{\bf ALBERT}}    
&base &General &--          & \ul{0.82} ($\pm$ .06)     &    \ul{0.80} ($\pm$ .03)         &      \ul{0.33} ($\pm$.00)      & \ul{0.88} ($\pm$ .03)        \\
&GP &General &Pre-trained      & 0.80 ($\pm$ .07)     &     0.75 ($\pm$ .04)       &      0.31  ($\pm$.03)      &  0.86 ($\pm$ .03)           \\
\hline
{\bf Baseline}&-- &-- &--                 & 0.72        &   0.64              &       0.27      &     0.77        \\
\hline
\end{tabular}
\caption{Performance metrics of document review tasks using different versions of  language models. For each task, the bold font shows the best performance among all models while the underlined number corresponds to the best performance among different versions of one model.}
\label{table_performance_results}
\end{table*}
As we see in Table \ref{table_performance_results}, the relative performance of different models depends on the task. However, this relative performance is more similar for the three sentence level tasks (i.e., passage retrieval, text similarity and sentiment analysis)  compared to the entity recognition task which is a token level one (see table \ref{table_LMs}). In these three sentence classification tasks, customizing the general BERT language model on the domain specific corpus improved the performance. The improvement is in average $8.5\%$ of the performance of the base version. In these three tasks, the highest performance of DistilBERT, RoBERTa, and ALBERT can respectively achieve $95\%$, $97\%$, and $95\%$ of the best performance of BERT in average. These models also seem to benefit from domain-specific language model customization in fewer cases. Moreover, by comparing GR and LR versions of BERT and DistilBERT, we see that when starting from scratch, using legal tokens marginally improves the performance compared to using the default general-domain tokens. However, it still does not beat the impact of using pre-trained weights considering the size of our corpus and amount of language model training that we have performed (10 epochs). Even extending the pre-trained model with only some legal tokens (see GP and HP versions), degrades the performance for most of the tasks. These observations are, however, aligned with Figure \ref{fig:lm_losses} and reflect the fact that more training time is required to get the same performance for training a model with larger number of tokens. For the entity recognition task, the base language models seem to perform better in general.

 \begin{itemize}[leftmargin=*,topsep=1pt,itemsep=1pt,partopsep=4pt, parsep=4pt]
\item {\bf How to choose a model if the computational resources are limited?}
\end{itemize}

\begin{table*}[h!]
{\begin{tabular}{|P{1.6cm}||P{1.35cm}|P{1.4cm}|P{1.7cm}|P{1.5cm}||P{1.35cm}|P{1.4cm}|P{1.7cm}|P{1.5cm}|}
\cline{2-9}
\multicolumn{1}{c|}{} &\multicolumn{4}{c||}{{\bf Training Time} (hours per epoch)}&\multicolumn{4}{c|}{{\bf Prediction Time} (milliseconds per sample)}\\
\hline
{\bf Model} & \shortstack{{\bf Passage} \\ {\bf Retrieval}} & \shortstack{{\bf Text} \\ {\bf Similarity}} & \shortstack{{\bf Entity} \\ {\bf Recognition}} & \shortstack{{\bf Sentiment} \\ {\bf Analysis}} & \shortstack{{\bf Passage} \\ {\bf Retrieval}} & \shortstack{{\bf Text} \\ {\bf Similarity}} & \shortstack{{\bf Entity} \\ {\bf Recognition}} & \shortstack{{\bf Sentiment} \\ {\bf Analysis}} \\

\hline
BERT&   1.53  & 0.44    &1.01 & 0.16&63&65&37&65\\
DistilBERT& 1.71    & 0.26    &0.35 &0.03&32&32&20&33\\
RoBERTa& 2.22    & 0.37 & 0.83 &0.43&60&63&35&63\\
ALBERT& 1.05    & 0.28 & 0.75 &0.66&69&68&22&71\\
\hline
\end{tabular}}
\caption{Training and prediction times for downstream document review tasks.}
\label{table_time_reports}
\end{table*}

The downstream performance metrics reflect the quality of the results we can get by using each of the models. However, there are other factors that should be taken into account in order to implement the models in a practical setup such as model size, memory usage, training and prediction times. In order to answer the third question, we report the training/inference time for all models. We have compared the language models based on their model size and number of training parameters in Table \ref{table_LMs}. This table also shows the time it takes to train each language model architecture for one epoch on the same GPU machine while fixing the batch size. Table \ref{table_time_reports} presents a summary of training and prediction time for fine-tuning the network for each of the downstream review tasks. The training and prediction times are reported for the model with the best performance for each task as reported in Table \ref{table_performance_results}. For each task, the GPU machine, learning rate, batch size, and the stopping criteria are the same when training different models. We see that training and prediction times are in general aligned with the number of parameters and layers reported  for different models in Table \ref{table_LMs}.

\section{Conclusion}\label{Conclusion}
In this paper, we investigated how different contextualized Transformer-based language models can be employed to automate different tasks of reviewing legal documents. We elaborated on the distinguishing features of the legal domain texts and studied several strategies for adapting the language models to the legal domain. Rather than using the standalone NLP tasks, we have compared the overall performance of models on real review scenarios. Our experiments show that while the token level task performs better with the general-domain pre-trained models, the sentence level tasks may benefit from some customization of language models. We also reported other practical aspects of models such as memory usage and training and prediction times. As a future work, we intend to investigate the performance of the same language models in a multi-task architecture where multiple document review tasks are co-trained.

\section{Acknowledgements}
We would like to thank Vector Institute\footnote{\url{https://vectorinstitute.ai/}} for the support during the  NLP industry collaborative project.

\bibliographystyle{IEEEtran}
\bibliography{IEEEexample}

% Generated by IEEEtran.bst, version: 1.12 (2007/01/11)
\begin{thebibliography}{10}
\providecommand{\url}[1]{#1}
\csname url@samestyle\endcsname
\providecommand{\newblock}{\relax}
\providecommand{\bibinfo}[2]{#2}
\providecommand{\BIBentrySTDinterwordspacing}{\spaceskip=0pt\relax}
\providecommand{\BIBentryALTinterwordstretchfactor}{4}
\providecommand{\BIBentryALTinterwordspacing}{\spaceskip=\fontdimen2\font plus
\BIBentryALTinterwordstretchfactor\fontdimen3\font minus
  \fontdimen4\font\relax}
\providecommand{\BIBforeignlanguage}[2]{{%
\expandafter\ifx\csname l@#1\endcsname\relax
\typeout{** WARNING: IEEEtran.bst: No hyphenation pattern has been}%
\typeout{** loaded for the language `#1'. Using the pattern for}%
\typeout{** the default language instead.}%
\else
\language=\csname l@#1\endcsname
\fi
#2}}
\providecommand{\BIBdecl}{\relax}
\BIBdecl

\bibitem{sulea2017exploring}
O.-M. Sulea, M.~Zampieri, S.~Malmasi, M.~Vela, L.~P. Dinu, and J.~Van~Genabith,
  ``Exploring the use of text classification in the legal domain,'' \emph{arXiv
  preprint arXiv:1710.09306}, 2017.

\bibitem{pan2009survey}
S.~J. Pan and Q.~Yang, ``A survey on transfer learning,'' \emph{IEEE
  Transactions on Knowledge and Data Engineering (TKDE)}, vol.~22, no.~10, pp.
  1345--1359, 2009.

\bibitem{jing2019survey}
K.~Jing, J.~Xu, and B.~He, ``A survey on neural network language models,''
  \emph{arXiv preprint arXiv:1906.03591}, 2019.

\bibitem{vaswani2017attention}
A.~Vaswani, N.~Shazeer, N.~Parmar, J.~Uszkoreit, L.~Jones, A.~N. Gomez,
  {\L}.~Kaiser, and I.~Polosukhin, ``Attention is all you need,'' in
  \emph{Advances in Neural Information rPocessing Systems (NeurIPS)}, 2017, pp.
  5998--6008.

\bibitem{elwany2019bert}
E.~Elwany, D.~Moore, and G.~Oberoi, ``Bert goes to law school: Quantifying the
  competitive advantage of access to large legal corpora in contract
  understanding,'' in \emph{Workshop on Document Intelligence at International
  Conference on Neural Information Processing Systems (NeurIPS)}, 2019.

\bibitem{chalkidis2020legal}
I.~Chalkidis, M.~Fergadiotis, P.~Malakasiotis, N.~Aletras, and
  I.~Androutsopoulos, ``Legal-bert: The muppets straight out of law school,''
  \emph{arXiv preprint arXiv:2010.02559}, 2020.

\bibitem{zhang2020rapid}
R.~Zhang, W.~Yang, L.~Lin, Z.~Tu, Y.~Xie, Z.~Fu, Y.~Xie, L.~Tan, K.~Xiong, and
  J.~Lin, ``Rapid adaptation of bert for information extraction on
  domain-specific business documents,'' \emph{arXiv preprint arXiv:2002.01861},
  2020.

\bibitem{devlin2019bert}
J.~Devlin, M.-W. Chang, K.~Lee, and K.~Toutanova, ``Bert: Pre-training of deep
  bidirectional transformers for language understanding,'' in \emph{Proceedings
  of the Annual Conference of the North American Chapter of the Association for
  Computational Linguistics: Human Language Technologies (NAACL-HLT)}, 2019.

\bibitem{peters2019tune}
M.~E. Peters, S.~Ruder, and N.~A. Smith, ``To tune or not to tune? adapting
  pretrained representations to diverse tasks,'' in \emph{Workshop on
  Representation Learning for NLP (RepL4NLP)}, 2019, pp. 7--14.

\bibitem{yang2019end}
W.~Yang, Y.~Xie, A.~Lin, X.~Li, L.~Tan, K.~Xiong, M.~Li, and J.~Lin,
  ``End-to-end open-domain question answering with bertserini,'' in
  \emph{Proceedings of the International Conference of the North American
  Chapter of the Association for Computational Linguistics (Demonstrations)},
  2019, pp. 72--77.

\bibitem{zhu2015aligning}
Y.~Zhu, R.~Kiros, R.~Zemel, R.~Salakhutdinov, R.~Urtasun, A.~Torralba, and
  S.~Fidler, ``Aligning books and movies: Towards story-like visual
  explanations by watching movies and reading books,'' in \emph{Proceedimgs of
  IEEE International Conference on Computer Vision (ICCV)}, 2015, pp. 19--27.

\bibitem{lee2020biobert}
J.~Lee, W.~Yoon, S.~Kim, D.~Kim, C.~So, and J.~Kang, ``Biobert: a pre-trained
  biomedical language representation model for biomedical text mining,''
  \emph{Bioinformatics (Oxford, England)}, vol.~36, no.~4, pp. 1234--1240,
  2020.

\bibitem{araci2019finbert}
D.~Araci, ``Finbert: Financial sentiment analysis with pre-trained language
  models,'' \emph{arXiv preprint arXiv:1908.10063}, 2019.

\bibitem{howard2018universal}
J.~Howard and S.~Ruder, ``Universal language model fine-tuning for text
  classification,'' in \emph{Proceedings of the Annual Meeting of the
  Association for Computational Linguistics (ACL)}, 2018, pp. 328--339.

\bibitem{peters2018deep}
M.~E. Peters, M.~Neumann, M.~Iyyer, M.~Gardner, C.~Clark, K.~Lee, and
  L.~Zettlemoyer, ``Deep contextualized word representations,'' in
  \emph{Proceedings of the Annual Conference of the North American Chapter of
  the Association for Computational Linguistics: Human Language Technologies
  (NAACL-HLT)}, 2018, pp. 2227--2237.

\bibitem{beltagy2019scibert}
I.~Beltagy, A.~Cohan, and K.~Lo, ``Scibert: Pretrained contextualized
  embeddings for scientific text,'' \emph{arXiv preprint arXiv:1903.10676},
  2019.

\bibitem{tan2015lstm}
M.~Tan, C.~d. Santos, B.~Xiang, and B.~Zhou, ``Lstm-based deep learning models
  for non-factoid answer selection,'' \emph{arXiv preprint arXiv:1511.04108},
  2015.

\bibitem{graves2013speech}
A.~Graves, A.-r. Mohamed, and G.~Hinton, ``Speech recognition with deep
  recurrent neural networks,'' in \emph{Proceedings of the IEEE International
  Conference on Acoustics, Speech and Signal Processing (ICASSP)}, 2013, pp.
  6645--6649.

\bibitem{chen2016xgboost}
T.~Chen and C.~Guestrin, ``Xgboost: A scalable tree boosting system,'' in
  \emph{Proceedings of the International Conference on Knowledge Discovery and
  Data Mining (ACM SIGKDD)}, 2016, pp. 785--794.

\bibitem{lai2018review}
T.~Lai, T.~Bui, and S.~Li, ``A review on deep learning techniques applied to
  answer selection,'' in \emph{Proceedings of the International Conference on
  Computational Linguistics (COLING)}, 2018, pp. 2132--2144.

\bibitem{alschner2019sense}
W.~Alschner, ``Sense and similarity: Automating legal text comparison,''
  \emph{Computational Legal Studies: The Promise and Challenge of Data-driven
  Research}, 2019.

\bibitem{neill2017classifying}
J.~O. Neill, P.~Buitelaar, C.~Robin, and L.~O. Brien, ``Classifying sentential
  modality in legal language: a use case in financial regulations, acts and
  directives,'' in \emph{Proceedings of the International Conference on
  Artificial Intelligence and Law (ICAIL)}, 2017, pp. 159--168.

\bibitem{verstraete2005scalar}
J.-C. Verstraete, ``Scalar quantity implicatures and the interpretation of
  modality: Problems in the deontic domain,'' \emph{Journal of pragmatics},
  vol.~37, no.~9, pp. 1401--1418, 2005.

\bibitem{wu2016google}
Y.~Wu, M.~Schuster, Z.~Chen, Q.~V. Le, M.~Norouzi, W.~Macherey, M.~Krikun,
  Y.~Cao, Q.~Gao, K.~Macherey \emph{et~al.}, ``Google's neural machine
  translation system: Bridging the gap between human and machine translation,''
  \emph{arXiv preprint arXiv:1609.08144}, 2016.

\bibitem{liu2019roberta}
Y.~Liu, M.~Ott, N.~Goyal, J.~Du, M.~Joshi, D.~Chen, O.~Levy, M.~Lewis,
  L.~Zettlemoyer, and V.~Stoyanov, ``Roberta: A robustly optimized bert
  pretraining approach,'' \emph{arXiv preprint arXiv:1907.11692}, 2019.

\bibitem{sanh2019distilbert}
V.~Sanh, L.~Debut, J.~Chaumond, and T.~Wolf, ``Distilbert, a distilled version
  of bert: smaller, faster, cheaper and lighter,'' \emph{arXiv preprint
  arXiv:1910.01108}, 2019.

\bibitem{wang2018glue}
A.~Wang, A.~Singh, J.~Michael, F.~Hill, O.~Levy, and S.~R. Bowman, ``Glue: A
  multi-task benchmark and analysis platform for natural language
  understanding,'' in \emph{Proceedings of the International Conference on
  Learning Representations (ICLR)}, 2018.

\bibitem{hinton2015distilling}
G.~Hinton, O.~Vinyals, and J.~Dean, ``Distilling the knowledge in a neural
  network,'' \emph{stat}, vol. 1050, p.~9, 2015.

\bibitem{bucilua2006model}
C.~Buciluǎ, R.~Caruana, and A.~Niculescu-Mizil, ``Model compression,'' in
  \emph{Proceedings of the International Conference on Knowledge Discovery and
  Data Mining (ACM SIGKDD)}, 2006, pp. 535--541.

\bibitem{lan2019albert}
Z.~Lan, M.~Chen, S.~Goodman, K.~Gimpel, P.~Sharma, and R.~Soricut, ``Albert: A
  lite bert for self-supervised learning of language representations,'' in
  \emph{Proceedings of the International Conference on Learning Representations
  (ICLR)}, 2019.

\bibitem{wolf2019huggingface}
T.~Wolf, L.~Debut, V.~Sanh, J.~Chaumond, C.~Delangue, A.~Moi, P.~Cistac,
  T.~Rault, R.~Louf, M.~Funtowicz \emph{et~al.}, ``Huggingface's transformers:
  State-of-the-art natural language processing,'' \emph{arXiv preprint
  arXiv:1910}, 2019.

\bibitem{guo2020wiki}
M.~Guo, Z.~Dai, D.~Vrande{\v{c}}i{\'c}, and R.~Al-Rfou, ``Wiki-40b:
  Multilingual language model dataset,'' in \emph{Proceedings of The 12th
  Language Resources and Evaluation Conference}, 2020, pp. 2440--2452.

\end{thebibliography}

\end{document}